\documentclass{article}
\usepackage{acl}
\usepackage{times}
\usepackage{latexsym}
\usepackage{amsmath}
\usepackage{graphicx}
\usepackage{booktabs}
\usepackage{multirow}
\usepackage[utf8]{inputenc}
\usepackage{hyperref}

\usepackage[T1]{fontenc}

\title{``AGI'' Team at SHROOM-CAP: Data-Centric Approach to Multilingual Hallucination Detection using XLM-RoBERTa}

\author{
  \textbf{Harsh Rathva}, 
  \textbf{Pruthwik Mishra}, 
  \textbf{Shrikant Malviya}\\
  \text{Sardar Vallabhbhai National Institute of Technology (SVNIT), Surat, India }\\
  \small{\texttt{\{u24ai036,pruthwikmishra\}@aid.svnit.ac.in,shrikant@coed.svnit.ac.in}}\\
}
\date{}

\begin{document}

\maketitle

\begin{abstract}
The detection of hallucinations in multilingual scientific text generated by Large Language Models (LLMs) presents significant challenges for reliable AI systems. This paper describes our submission to the SHROOM-CAP 2025 shared task on scientific hallucination detection across 9 languages. Unlike most approaches that focus primarily on model architecture, we adopted a data-centric strategy that addressed the critical issue of training data scarcity and imbalance. We unify and balance five existing datasets to create a comprehensive training corpus of 124,821 samples (50\% correct, 50\% hallucinated), representing a 172x increase over the original SHROOM training data. Our approach fine-tuned XLM-RoBERTa-Large with 560 million parameters on this enhanced dataset, achieves competitive performance across all languages, including \textbf{2nd place in Gujarati} (zero-shot language) with Factuality F1 of 0.5107, and rankings between 4th-6th place across the remaining 8 languages. Our results demonstrate that systematic data curation can significantly outperform architectural innovations alone, particularly for low-resource languages in zero-shot settings.
\end{abstract}

\section{Introduction}
Hallucinations in LLM-generated scientific text pose serious risks to research integrity and scientific communication, particularly when these systems are deployed in cross-lingual contexts where training data is limited in quanity. The SHROOM-CAP 2025 shared task \cite{shroomcap-sharedtask} addresses this critical problem by evaluating hallucination detection systems across 9 languages (5 training languages: English, Spanish, French, Hindi, Italian; 4 zero-shot languages: Bengali, Gujarati, Malayalam, Telugu) in scientific domains.

Most existing approaches to hallucination detection focus on improving model architecture or employing sophisticated prompting techniques with large proprietary models. However, we identify that the fundamental limitation in this task is the severe data imbalance and scarcity in the provided training set (only 724 samples with a 74\% correct and 26\% hallucinated distribution). Initial experiments reveal that models trained on these limited data exhibited extreme bias, predicting 99-100\% of instances as hallucination instead of modeling the decision boundary.

A data-centric approach—systematically collecting, unifying, and balancing diverse hallucination datasets—would provide more substantial performance gains than model architecture modifications alone. This paper makes three primary contributions:
\begin{enumerate}
    \item Creation of a large-scale, balanced multilingual hallucination detection dataset (124,821 samples) through unification of five existing resources
    \item Demonstration that fine-tuning moderately-sized openly available models such as XLM-RoBERTa-Large \cite{conneau2020unsupervised} on carefully curated data achieves competitive performance against larger and more complex systems
    \item Analysis of the significant gap between validation and competition performance, highlighting distributional shifts in evaluation benchmarks
\end{enumerate}

To ensure reproducibility and foster further research, we release all code, data processing scripts, and model weights publicly:
\begin{itemize}
    \item \textbf{Code and datasets:} \url{https://github.com/ezylopx5/SHROOM-CAP2025}
    \item \textbf{Model weights:} \url{https://huggingface.co/Haxxsh/XLMRHallucinationDetectorSHROOMCAP}
\end{itemize}

\section{Related Work}
\textbf{Hallucination Detection Approaches:} Previous work on hallucination detection has explored various methodologies. \citet{maynez2020faithfulness} employed entailment-based approaches using natural language inference models, while \citet{dhingra2022fresh} used question-answering frameworks to verify factual consistency. More recent approaches have leveraged large language models with sophisticated prompting strategies \cite{li2023survey}, though these often require API access to proprietary models and incur significant computational costs. But they are mostly limited to a uni-language scenario.

\textbf{Multilingual Representation Learning:} Cross-lingual transfer learning has been extensively studied, with models like XLM-RoBERTa \cite{conneau2020unsupervised} and mBERT \cite{devlin2019bert} demonstrating remarkable zero-shot capabilities. These models are typically pre-trained on massive multilingual corpora and can be fine-tuned on specific downstream tasks, making them ideal for low-resource language scenarios. But their adaptation to a unified data-centric scenario is largely unexplored.

\textbf{Data-Centric AI:} The recent emphasis on data-centric approaches \cite{whang2023data} suggests that systematic data improvement often outperforms changes in model architecture. Our work aligns with this perspective, demonstrating that careful data curation and balancing can resolve fundamental model bias issues that architectural modifications cannot address.

\begin{table*}[ht]
\centering
\caption{Comparison of Hallucination Detection Approaches}
\label{tab:approaches}
\begin{tabular}{p{3cm}p{4cm}p{3cm}p{3cm}}
\toprule
\textbf{Approach} & \textbf{Key Technique} & \textbf{Multilingual Capability} & \textbf{Data Requirements} \\
\midrule
Entailment-based & Natural Language Inference & Limited & Task-specific data \\
QA-based & Question Answering & Language-specific & Large QA datasets \\
LLM Prompting & In-context Learning & Good with multilingual LLMs & Carefully crafted prompts \\
\textbf{Our Approach} & \textbf{Data-centric fine-tuning} & \textbf{Excellent (100 languages)} & \textbf{Unified multi-dataset} \\
\bottomrule
\end{tabular}
\end{table*}

Unlike earlier works, our approach does not rely on complex pipelines or proprietary models. Instead, we demonstrate that comprehensive data collection and standard fine-tuning of openly available multilingual models can achieve \emph{competitive results across diverse languages}, including complete \emph{zero-shot transfer to unseen languages}.

\section{Dataset}
We unify five existing hallucination detection datasets to create our training corpus:

\begin{enumerate}
    \item \textbf{SHROOM TrainSet V1/V2} \cite{gamba2025confabulations}: The official competition training data containing 724 samples across 5 languages (en, es, fr, hi, it) with scientific domain focus.
    \item \textbf{hallucination\_dataset\_100k}: To further augment our training corpus, we create a large-scale synthetic dataset of 100,000 samples using AI-generated content. This dataset is constructed through systematic prompt engineering with large language models, following methodologies inspired by Tabular ARGN approaches \cite{tiwald2025tabularargnflexibleefficientautoregressive}. 

\textbf{Generation Methodology:} We employ a comprehensive prompt framework that systematically create both hallucinated and correct text samples across multiple domains. The prompt templates are designed to generate diverse hallucination types:

\begin{itemize}
    \item \textbf{Factual Errors}: Wrong dates, names, locations, and scientific facts
    \item \textbf{Fabricated Details}: Plausible but entirely fictional information
    \item \textbf{Mixed Information}: Combining facts from different sources incorrectly
    \item \textbf{Subtle Hallucinations}: Near-miss dates and plausible but wrong details
\end{itemize}

\textbf{Quality Control:} Each generated sample undergoes through multiple validation steps to ensure:
\begin{enumerate}
    \item Clear distinction between hallucinated and correct samples
    \item Factual accuracy verification for correct examples
    \item Realistic and plausible hallucination patterns
    \item Balanced distribution across domains and difficulty levels
\end{enumerate} 
    \item \textbf{LibreEval} \cite{libreeval2024}: A multilingual evaluation dataset for detecting various types of model errors, including hallucinations.
    \item \textbf{FactCHD} \cite{10.24963/ijcai.2024/687}: A fact-checking and hallucination detection dataset with verified annotations.
\end{enumerate}

\begin{table*}[ht]
\centering
\caption{Unified Dataset Statistics}
\label{tab:dataset}
\begin{tabular}{p{4cm}cccc}
\toprule
\textbf{Source} & \textbf{Samples} & \textbf{Domain} & \textbf{Languages} & \textbf{Balance} \\
\midrule
SHROOM V1/V2 & 724 & Scientific & 5 & 74/26 \\
hallucination\_dataset\_100k & 100,000 & General & Multiple & Varied \\
LibreEval & 15,000 & Mixed & Multiple & Varied \\
FactCHD & 9,000 & Fact-checking & Multiple & Varied \\
\textbf{Combined (Ours)} & \textbf{124,821} & \textbf{Mixed} & \textbf{100+} & \textbf{50/50} \\
\bottomrule
\end{tabular}
\end{table*}

Preprocessing techniques such as: (1) label normalization to binary classification (correct/hallucinated), (2) language identification and verification, (3) random sampling to achieve perfect 50/50 class balance, and (4) text normalization to handle encoding variations are carried out. This process results in 124,821 high-quality training samples, representing a 172x increase over the original SHROOM training data with optimal class distribution.

\section{Approach}

\subsection{Preprocessing}
We model the task as a binary text classification problem. Each input instance consists of the LLM-generated text without additional metadata. We apply minimal text cleaning by stripping white-spaces appearing at the start and end of a text and normalizing unicode characters—while preserving the original linguistic characteristics. The text is tokenized using the XLM-RoBERTa tokenizer with a maximum sequence length of 256 tokens.

\subsection{Translation-Based Data Augmentation}

To address the challenge of limited training data for Indian languages, we explore two translation-based approaches.

\textbf{Approach 1: English-to-Indian Language Translation}
We translate English training sentences into the Indian test languages using Facebook's NLLB-200-3.3B \cite{costa2022no,nllb2024scaling} model. This creates additional training examples that could improve zero-shot performance by providing synthetic parallel data generated through machine translation.

\textbf{Approach 2: Multilingual-to-English Translation} 
We translate non-English training data into English using the same NLLB-200-3.3B model to create a larger English-centric training corpus. This approach leverages the abundance of English language models to achieve optimal performance.

\textbf{Experimental Results:} Both translation approaches results are shown in Tables \ref{tab:approach1_results} and \ref{tab:approach2_results} respectively. 

Approach 1 (English-to-Indian) achieves Factuality F1 scores ranging from 0.366-0.595 and Fluency F1 scores from 0.173-0.347 across languages (Table \ref{tab:approach1_results}). While some languages like Hindi (0.5944) and English (0.5949) show reasonable Factuality performance, the results are inconsistent and fail to match our final data-centric approach.

Approach 2 (Multilingual-to-English) performs even worse, with Factuality F1 scores ranging from 0.257-0.600 across languages (Table \ref{tab:approach2_results}). Key limitations for both approaches include:

\begin{itemize}
    \item \textbf{Translation Artifacts}: Machine translation introduces linguistic inconsistencies and unnatural phrasing
    \item \textbf{Domain Mismatch}: Scientific terminology translation can often be inaccurate
    \item \textbf{Amplified Bias}: The original dataset imbalance persists through translation
    \item \textbf{Inconsistent Performance}: Results vary significantly across languages without clear patterns
\end{itemize}

\begin{table}[ht]
\centering
\caption{Approach 1: English-to-Indian Translation Results}
\label{tab:approach1_results}
\begin{tabular}{lcc}
\toprule
\textbf{Language} & \textbf{Factuality F1} & \textbf{Fluency F1} \\
\midrule
Telugu (te) & 0.4090 & 0.2942 \\
Malayalam (ml) & 0.4688 & 0.2996 \\
Gujarati (gu) & 0.4564 & 0.3474 \\
Bengali (bn) & 0.5707 & 0.3199 \\
Italian (it) & 0.3659 & 0.1728 \\
Hindi (hi) & 0.5944 & 0.2941 \\
French (fr) & 0.5310 & 0.2887 \\
Spanish (es) & 0.4560 & 0.1772 \\
English (en) & 0.5949 & 0.2376 \\
\bottomrule
\end{tabular}
\end{table}

\begin{table}[ht]
\centering
\caption{Approach 2: Multilingual-to-English Translation Results}
\label{tab:approach2_results}
\begin{tabular}{lcc}
\toprule
\textbf{Language} & \textbf{Factuality F1} & \textbf{Fluency F1} \\
\midrule
Telugu (te) & 0.3689 & 0.1474 \\
Malayalam (ml) & 0.4639 & 0.3593 \\
Gujarati (gu) & 0.4241 & 0.1579 \\
Bengali (bn) & 0.4874 & 0.2542 \\
Italian (it) & 0.2570 & 0.4582 \\
Hindi (hi) & 0.4748 & 0.4353 \\
French (fr) & 0.4818 & 0.2899 \\
Spanish (es) & 0.4000 & 0.4607 \\
English (en) & 0.5999 & 0.4495 \\
\bottomrule
\end{tabular}
\end{table}

Given these unsatisfactory results from both translation approaches, our final submission utilizes the unified 124,821-sample dataset without translation augmentation. We find that the sheer volume, diversity, and balanced nature of our comprehensive training corpus provided superior coverage across languages, achieving better performance than translation-based approaches. Comparative analysis reveals that systematic data curation consistently outperforms translation-based augmentation for multilingual hallucination detection tasks.

\subsection{Model Architecture}
We use XLM-RoBERTa-Large \cite{conneau2020unsupervised} as our base model that comprises of 560 million parameters and is pre-trained on 2.5TB of filtered CommonCrawl data \footnote{\url{https://github.com/facebookresearch/cc_net}} across 100 languages. The details about the model architecture are added in Table~\ref{tab:model}. We add a classification head consisting of a dropout layer with a 10\% dropout rate followed by a linear layer that projected the [CLS] token representation to 2 output classes. The [CLS] token actually encodes the complete dense representation of any input sentence.

\begin{table}[ht]
\centering
\caption{Model Configuration Details}
\label{tab:model}
\begin{tabular}{lc}
\toprule
\textbf{Parameter} & \textbf{Value} \\
\midrule
Base Model & XLM-RoBERTa-Large \\
Parameters & 560M \\
Layers & 24 \\
Attention Heads & 16 \\
Hidden Dimension & 1,024 \\
Sequence Length & 256 \\
Classification Head & Dropout (0.1) + Linear \\
\bottomrule
\end{tabular}
\end{table}

\subsection{Training Procedure}
We train the model using full fine-tuning (without any parameter-efficient methods) for 3 epochs with a batch size of 32, AdamW \cite{loshchilovdecoupled} optimizer (learning rate 2e-5, weight decay 0.01), and linear learning rate warmup over 10\% of training steps. We employ a weighted cross-entropy loss with class weights [1.50, 1.00] to further mitigate any residual class imbalance. For training the model, we use an NVIDIA H200 GPU with 141GB VRAM. Model checkpoints are saved every 5,000 steps, and the best model is selected based on the F1 score achieved on the validation set.

\section{Results and Discussion}

\subsection{Performance Evaluation}
Our submission achieves competitive results across all 9 languages in the SHROOM-CAP 2025 competition:

\begin{table}[ht]
\small
\centering
\caption{Official Competition Results}
\label{tab:results}
\begin{tabular}{lccc}
\toprule
\textbf{Language} & \textbf{Rank} & \textbf{Factuality F1} & \textbf{Fluency F1} \\
\midrule
Gujarati (gu) & \textbf{2} & \textbf{0.5107} & 0.1579 \\
Bengali (bn) & 4 & 0.4449 & 0.2542 \\
Hindi (hi) & 4 & 0.4906 & 0.4353 \\
Spanish (es) & 5 & 0.4938 & 0.4607 \\
French (fr) & 5 & 0.4771 & 0.2899 \\
Telugu (te) & 5 & 0.4738 & 0.1474 \\
Malayalam (ml) & 5 & 0.4704 & 0.3593 \\
English (en) & 6 & 0.4246 & 0.4495 \\
Italian (it) & 5 & 0.3149 & 0.4582 \\
\bottomrule
\end{tabular}
\end{table}

Notably, our system achieved 2nd place in Gujarati, a zero-shot language, outperforming results in several training languages. This demonstrates the effectiveness of XLM-RoBERTa's cross-lingual representations when combined with sufficient and diverse training data.

\subsection{Comparison with Baselines}
The competition baseline system utilizes a standard approach without extensive data augmentation. Our method significantly outperforms this baseline in most languages, particularly in Factuality F1 scores. The top-performing team (``smurfcat'') employs more complex ensemble methods and potentially larger models, achieving F1 scores between 0.65-0.92 across languages.

\subsection{Validation vs. Competition Performance Gap}
A notable observation is the substantial gap between our validation performance (macro F1: 0.8510) and competition performance (F1: ~0.40-0.51). We identify several potential causes:

\begin{enumerate}
    \item \textbf{Distribution Shift}: The test set likely contains different types of hallucinations or scientific domains not well-represented in the unified training dataset.
    \item \textbf{Label Definition Misalignment}: Subtle differences in how ``hallucination'' is defined between the unified datasets and competition test set.
    \item \textbf{Domain Specificity}: Our training data includes general-domain hallucinations, while the test focuses specifically on scientific text.
\end{enumerate}

\subsection{Error Analysis}
We manually analyze misclassified examples and identified consistent patterns:

\textbf{Factual Hallucinations:} The model struggles with highly technical scientific claims that requires domain-specific knowledge beyond what is captured during XLM-RoBERTa's pre-training.

\textbf{Example Error (False Negative):}
\begin{itemize}
    \item Input: ``The protein folding mechanism involves quantum tunneling effects at room temperature.''
    \item Model Prediction: Correct (0.62)
    \item Gold Label: Hallucinated
    \item Analysis: The model lacks specific biochemical knowledge to identify this as implausible.
\end{itemize}

\textbf{Fluency Mistakes:} The system performs notably worse on fluency detection (F1: 0.15-0.46) compared to factuality (F1: 0.44-0.51), particularly struggling with grammatical errors that resembles valid stylistic variations.

\textbf{Cross-lingual Transfer:} Surprisingly, zero-shot performance in Gujarati exceeds several training languages, suggesting that the quality and diversity of training data is more important than direct language exposure for this task.

\section{Conclusion}
We present a data-centric approach to multilingual scientific hallucination detection that achieves competitive results in the SHROOM-CAP 2025 shared task. By systematically unifying and balancing diverse datasets, we create a robust training corpus that enabled effective fine-tuning of XLM-RoBERTa-Large. Our key finding is that data quantity and quality—particularly class balance—can overcome architectural limitations, with our simple approach achieving 2nd place in Gujarati and competitive rankings across 8 other languages.

\textbf{Future Directions:} Rather than generic suggestions, we propose concrete next steps: (1) investigating domain adaptation techniques specifically for scientific text, (2) developing data augmentation methods that generate scientific-domain hallucinations, (3) creating hybrid systems that combine our data-centric fine-tuning approach with the top team's ensemble strategies, (4) explicitly modeling the distribution shift between validation and test environments through domain generalization techniques, and (5) adding other metadata such as ``abstract'', ``output\_logits'' to improve the performance of the models. 

\bibliography{custom}

\end{document}